# Geometric Graph Representations and Geometric Graph Convolutions for Deep Learning on Three-Dimensional (3D) Graphs

Daniel T. Chang (张遵)

*IBM (Retired)* dtchang43@gmail.com

**Abstract:**

The geometry of three-dimensional (3D) graphs, consisting of nodes and edges, plays a crucial role in many important applications. An excellent example is molecular graphs, whose geometry influences important properties of a molecule including its reactivity and biological activity. To facilitate the incorporation of geometry in deep learning on 3D graphs, we define three types of geometric graph representations: positional, angle-geometric and distance-geometric. For proof of concept, we use the distance-geometric graph representation for geometric graph convolutions. Further, to utilize standard graph convolution networks, we employ a simple edge weight / edge distance correlation scheme, whose parameters can be fixed using reference values or determined through Bayesian hyperparameter optimization. The results of geometric graph convolutions, for the ESOL and Freesol datasets, show significant improvement over those of standard graph convolutions. Our work demonstrates the feasibility and promise of incorporating geometry, using the distance-geometric graph representation, in deep learning on 3D graphs.

## 1 Introduction

The geometry of *three-dimensional (3D) graphs*, consisting of nodes and edges, plays a crucial role in many important applications. An excellent example is *molecular graphs*, whose geometry influences important properties of a molecule including its reactivity and biological activity.

The *geometry* of 3D graphs is often specified in terms of the *Cartesian coordinates of nodes*. However, such specification depends on the (arbitrary) choice of origin and is too general for specifying geometry. We focus on 3D graphs whose geometry can be fully specified in terms of *edge distances (d), angles (θ)* and *dihedrals (φ)*. A key advantage of such specification is its *invariance to rotation and translation* of the graph.

*Distance geometry* [15] is the characterization and study of the geometry of 3D graphs based only on given values of the *distances* between pairs of nodes. From the perspective of distance geometry, the geometry of 3D graphs can be equivalently specified in terms of *edge distances (d), angle distances ($d^θ$)* and *dihedral distances ($d^φ$)*. In addition to *invariance to rotation and translation* of the graph, such specification adopts a *unified scheme (distance)* and reflects *pair-wise node interactions* and their generally local nature, which are additional advantages.

To facilitate the incorporation of geometry in deep learning on 3D graphs, we define three types of *geometric graph representations:* positional, angle-geometric and distance-geometric. The *positional graph representation* is based on node positions, i.e., Cartesian coordinates of nodes. The *angle-geometric graph representation* centers on edge distances, angles and dihedrals; it is invariant to rotation and translation of the graph. The *distance-geometric graph representation* is based on distances: edge distances, angle distances and dihedral distances; it is invariant to rotation and translation of the graph and it reflects pair-wise node interactions and their generally local nature.

For proof of concept, we use the distance-geometric graph representation for *geometric graph convolutions*. Further, to utilize standard *graph convolutional networks (GCNs)* [4] for geometric graph convolutions, we employ a simple *edge weight / edge distance correlations* scheme, whose parameters can be fixed using *reference values* or determined using *Bayesian hyperparameter optimization* [2].

GCNs have been applied to deep learning on graphs. However, *standard graph convolutions* do not take spatial arrangements of the nodes and edges into account. Therefore, they can accommodate only graph constitution, but not graph geometry. Recently, there have been efforts to extend GCNs by incorporating *3D node coordinates* in graph convolutions [5-6]. The proposed schemes, however, only consider *adjacent nodes*. They do not consider other nodes that are important to graph geometry. These include *second-neighbor nodes* which are part of the edges that form angles and *third-neighbor nodes* which are part of the edges that form dihedrals.

The combination of using the distance-geometric graph representation and employing an edge weight / edge distance correlations scheme enables us to incorporate the full geometry of 3D graphs in graph convolutions utilizing standard GCNs by (1) expanding the kinds of edges involved to include not just *edges* with neighbor nodes, but also *angle edges* with second-neighbor nodes and *dihedral edges* with third-neighbor nodes and (2) assigning different *weight*s to different edges based on their kind and their distance.

For molecular graphs, edge distance corresponds to *bond length* and edge weight to *bond strength*. Bond strength is empirically related to bond length through power laws with parameters $R_0$ and N [11-12]. We leverage this knowledge and correlate edge weight with edge length through *power laws with parameters $R_0$ and N* in the feasibility study, which is on molecular graphs. However, this is not necessary in general. For other types of 3D graphs, even for molecular graphs, other forms of edge weight / edge distance correlations can be used, e.g., exponential functions.

For the feasibility study, we implemented geometric graph convolutions using *PyTorch Geometric (PyG)* [1, 3] and Bayesian hyperparameter optimization using *BoTorch, GPyTorch* and *Ax* [2, 24-30]. We used the *ESOL* and *FreeSolv* datasets provided by geo-GCN [6], which contain molecular graph data including three-dimensional node coordinates.

## 2 Geometry of 3D Graphs

The *geometry* of 3D graphs is the three-dimensional arrangement of the *nodes* and *edges* in a graph. It is often specified in terms of the *Cartesian coordinates of nodes*. However, such specification depends on the (arbitrary) choice of origin and is too general for specifying geometry.

We focus on 3D graphs whose geometry can be fully specified in terms of *edge distances (d), angles (θ)* and *dihedrals (φ)*. Molecular graphs are excellent examples of such graphs. The edge distance is the distance between two nodes connected together. The angle is the angle formed between three nodes across two edges. For three edges connected in a chain, the dihedral is the angle between the plane formed by the first two edges and the plane formed by the last two edges. These are illustrated in the following diagram:

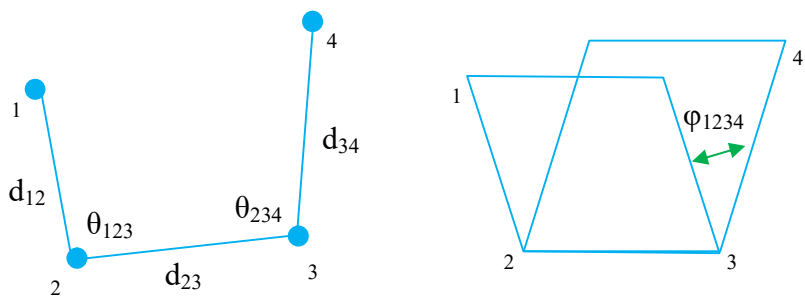

A key advantage of using edge distances, angles and dihedrals to specify geometry is its *invariance to rotation and translation* of the graph.

### 2.1 Molecular Geometry

*Molecular geometry* (https://en.wikipedia.org/wiki/Molecular_geometry) is the three-dimensional arrangement of the *atoms* (nodes) and *bonds* (edges) in a molecule. It influences important properties of a molecule including its reactivity and biological activity. An excellent example is *3D-QSAR* [7-8] which exploits the three-dimensional properties of the molecules (ligands) to predict their biological activities. Chemical compounds containing one or more single bonds exist at each moment in many different *conformations*. Therefore, it is necessary to include various such conformations (geometries) of the molecules in a QSAR (Quantitative Structure Activity Relationship) study.

Molecular geometry can be specified in terms of *bond lengths* (i.e., edge distances), *bond angles* (i.e., angles) and *dihedral angles* (i.e., dihedrals). The bond length is the average distance between two atoms bonded together. The bond angle is the angle formed between three atoms across two bonds. For three bonds in a chain, the dihedral angle is the angle between the plane formed by the first two bonds and the plane formed by the last two bonds. These are no different from the general case of 3D graphs. Bond lengths, bond angles and dihedral angles can be calculated from the *Cartesian coordinates of atoms* in a molecule, which are generally expressed in the unit of *angstrom (Å)*.

*Bond lengths* (https://en.wikipedia.org/wiki/Bond_length) of carbon with other elements are in the range of *1 to 3 Å*. Bond lengths in organic compounds generally range from 1.08 to 1.54 Å. However, the existence of a very long C–C bond length of up to 3.05 Å has been found in dimers. The longest covalent bond is the bismuth-iodine single bond with length 2.81 Å. The bond length between a given pair of atoms may vary between different molecules.

Bond length is related to *bond order*, or *bond type* (single, double, triple, aromatic), and the correlation is decreasing and bend in conjugated hydrocarbons [9]. Bond lengths have been used for automatic bond perception [10], whose aim is to identify bond types, via decision-trees based machine learning. Bond order can be used to represent *bond strength*. However, there is no unique definition of bond strength. Other bond properties have been associated with bond strength, e.g., bond dissociation energy, the force constant of the bond, and the interatomic electron density of the bond.

*Bond strength* is inversely related to bond length: all other factors being equal, a stronger bond will be shorter. For oxides, the relationship is of the form $s = (R/R_o)^{-N}$ [11] where s = bond strength, R = bond length and $R_o$ and N are fitted parameters, with $R_o$ in the range of 1.1 to 2.9 Å and N from 2.2 to 6.0. The (inverse) relationship can also be expressed by the power law $R/R_0 = p^{-n}$ [12] where p = bond strength, with $R_o$ = 1.39 Å and n = 0.22 (thus n ≈ 1/N). In both cases, ln(s), or ln(p), has a linear dependence on ln(R). It should be noted that these correlations are empirical and they are not fundamental rules or laws [13]. (We will utilize the above information in Section 5.3.)

Predicting an accurate 3D molecular geometry is a crucial task for cheminformatics. Various methods are available including: RDKit (based on ETKDG, See Section 3.1), OpenBabel and fragment-based. These are discussed and compared in [14]. The outcomes of prediction include coordinates, bond lengths, bond angles and dihedral angles, among other things.

The *RDKit* (https://www.rdkit.org/) provides the method *GetBondLength()* to get the bond length in Å between bonded atoms i and j. It also provides the method *GetAngleDegree()* and *GetAngleRad()* to get the bond angle between bonded atoms i, j and k, as well as *GetDihedralDegree()* and *GetDihedralRad()* to get the dihedral angle between bonded atoms i, j, k and l.

## 3 Distance Geometry of 3D Graphs

*Distance geometry* [15] refers to a foundation of geometry based on the concept of *distances* instead of those of points and lines or point coordinates. In general, the focus of distance geometry is on the so-called *Distance Geometry Problem (DGP)* which can be stated as: given a weighted graph G = (V, E) and the dimension K of a vector space, draw G in $R^K$ such that each edge is drawn as a straight segment of distance equal to its weight. In other words, the essence of DGP is of reconstructing node positions from given edge distances. A DGP instance may have no solutions if the given distances do not define a metric, a finite number of solutions if the graph is rigid, or an unlimited number of solutions if the graph is flexible.

For 3D graphs, distance geometry is the characterization and study of their geometry based only on given values of the *distances* between pairs of nodes. From the perspective of distance geometry, therefore, the geometry of 3D graphs can be equivalently specified in terms of *edge distances (d), angle distances ($d^\theta$)* and *dihedral distances ($d^\varphi$)*. The angle distance is the distance of the *angle edge ($e^\theta$)* and the dihedral distance is the distance of the *dihedral edge ($e^\varphi$)*. The angle edge is the unconnected, end edge between the end nodes of an angle and the dihedral edge is the unconnected, end edge between the end nodes of a dihedral. (We therefore refer to both angle edges and dihedral edges as *end edges*.) These are illustrated in the following diagram (with dashed lines representing angle edges and dotted lines representing dihedral edges):

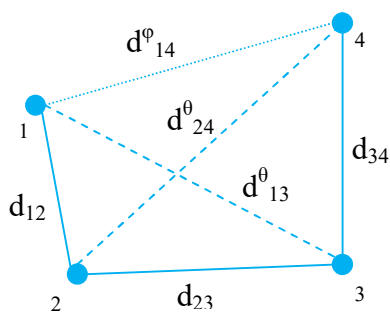

As the case of using edge distances, angles and dihedrals to specify geometry, a key advantage of specifying geometry in terms of edge distances, angle distances and dihedral distances is its *invariance to rotation and translation*. In addition, it adopts a *unified scheme (distance)* and reflects *pair-wise node interactions* and their generally local nature, which are additional advantages. These are very useful for *graph convolutions*, which are locally oriented. They are particular useful for *molecular graphs*, since electrostatic, intermolecular, and other conformation-driven properties of molecules depend on the pair-wise interatomic (internodal) distances.

## 3.1 Molecular Distance Geometry

Distance geometry is the basis for a *geometric theory of molecular conformation* [16]. A distance geometry description of a 3D molecular graph (conformation) consists of a list of *distance and chirality constraints*. These are, respectively, lower and upper bounds on the distances, and the chirality of selected quadruples of nodes. The distance geometry approach is predicated on the assumption that it is possible to adequately describe the set of all possible conformations, i.e., the *conformation space*, by means of such purely geometric descriptions.

In general, one tries to find a number of different solutions (i.e., conformations) that satisfy distance geometry descriptions. Such a set of conformations is called a *conformational ensemble*. There are many algorithms which together provide a means of generating conformational ensembles consistent with distance geometry descriptions. The *overall procedure* consists of the following steps [16]:

(1) Bound smoothing: Extrapolating a complete set of lower and upper limits on all the distances from the sparse set of lower and upper bounds that are usually available.
(2) Embedding: Choosing a random distance matrix from within these limits, and computing coordinates that are a certain best fit to the distances. It is critical that good *search methods* are used to direct search towards favorable regions of the conformational space.
(3) Optimization: Optimizing these coordinates versus an *'error' function* which measures the total violation of the distance and chirality constraints.

The conformation generator *ETKDG* [17] is a stochastic search method that utilizes distance geometry together with knowledge derived from experimental crystal structures. It is used in *RDKit* by default for conformation generation. It has been shown to generate good conformations for acyclic molecules. Recently it has been improved for conformation generation of molecules containing small or large aliphatic rings.

The *upper and lower bounds* on the distances are traditionally obtained from chemical knowledge and experimental data. However, deep learning approaches have begun to emerge. A *probabilistic generative model* is used in [18] to learn distributions, and therefore upper and lower bounds, of the interatomic distances of molecules from their *graph representations*. Specifically, it uses a variational autoencoder consisting of an inference model $q_\lambda(z|d,G)$ and a generative model $p_\gamma(d|z,G)p_\gamma(z|G)$. The graph representations are based on an *extended molecular graph* which consists of *nodes* (atoms), *edges* (bonds) and *auxiliary edges* (representing angles and dihedral angles). (Auxiliary edges are the same as *end edges* discussed in Section 3.) For unseen molecules, upon generation of the upper and lower bounds of the distances using the generative model, it uses a Euclidean distance geometry algorithm to generate conformations.

The *Conf17* benchmark is used in [18]. It is the first benchmark for conformation sampling and is based on the ISO17 dataset which consists of conformations of various molecules with the atomic composition $C_7H_{10}O_2$. For the Conf17 benchmark, *edge distances* range from 0.9 to 1.7 Å, *angle distances* from 1.5 to 3 Å, and *dihedral distances* from 2 to 4 Å. (We will utilize this information in Section 5.3.)

## 4 Geometric Graph Representations

To facilitate the incorporation of geometry in deep learning on 3D graphs, we define three types of *geometric graph representations:* positional, angle-geometric and distance-geometric, as discussed in the following.

### 4.1 Positional Graph Representation

In general, a 3D graph can be represented as $G = (X, (I, E), P)$ where $X \in R^{N \times d}$ is the *node feature matrix*, $(I, E)$ is the *sparse adjacency tuple*, and $P \in R^{N \times 3}$ is the *node position matrix*. $I \in N^{2 \times U}$ encodes *edge indices* in coordinate (COO) format and $E \in R^{U \times s}$ is the *edge feature matrix*. $P$ encodes the Cartesian coordinates of nodes. (N is the number of nodes, U is the number of edges, d is the number of node features, and s is the number of edge features.) This representation is based on node positions; therefore we refer to it as the *positional graph representation*. As discussed in Section 2, it depends on the (arbitrary) choice of the origin for the coordinates and it is too general for representing geometry.

### 4.2 Angle-Geometric Graph Representation

From the perspective of geometry, a 3D graph can be represented as $G = (X, (I, E), (D, \Theta, \Phi))$ where $D \in R^{U \times 1}$ is the *edge distance matrix*, $\Theta \in R^{U^\theta \times 1}$ is the *angle matrix*, and $\Phi \in R^{U^\varphi \times 1}$ is the *dihedral matrix*. (U is the number of edges, $U^\theta$ is the number of angles, and $U^\varphi$ is the number of dihedrals.) This representation centers on (edge distances,) angles and dihedrals;

therefore we refer to it as the *angle-geometric graph representation*. As discussed in Section 2, it is *invariant to rotation and translation* of the graph, which is a major advantage over the positional graph representation.

### 4.3 Distance-Geometric Graph Representation

From the perspective of distance geometry, a 3D graph can be represented as G = (**X**, (**I**, **E**), (**D**, **D**$^\theta$, **D**$^\varphi$)) where **D** ∈ R$^{U\times 1}$ is the *edge distance matrix*, **D**$^\theta$ ∈ R$^{U\theta \times 1}$ is the *angle distance matrix*, and **D**$^\varphi$ ∈ R$^{U\varphi \times 1}$ is the *dihedral distance matrix*. (U is the number of edges, U$^\theta$ is the number of angles, and U$^\varphi$ is the number of dihedrals.) This representation is based on distances; therefore we refer to it as the *distance-geometric graph representation*. It is *invariant to rotation and translation* of the graph, same as the angle-geometric graph representation. In addition, as discussed in Section 3, it reflects *pair-wise node interactions* and their generally local nature, which is another major advantage.

## 5 Geometric Graph Convolutions

*Geometric graph convolutions* extend standard graph convolutions [4] by incorporating the *3D geometry* of graphs in the graph convolution process. *Standard graph convolutions* do not take spatial arrangements of the nodes and edges into account. Therefore, they can accommodate only graph constitution, but not graph geometry.

Recently, there have been efforts to extend GCNs by incorporating *3D node coordinates* in graph convolutions [5-6]. The proposed schemes, however, only consider *adjacent nodes*. They do not consider other nodes that are important to graph geometry. These include *second-neighbor nodes* which are part of the edges that form angles and *third-neighbor nodes* which are part of the edges that form dihedrals.

### 5.1 Graph Convolutional Networks (GCNs)

As in [1], we use *graph neural networks* (*GNNs*) that employ the following *message passing* scheme for node i at layer k:

$$\mathbf{x}_i^{(k)} = \gamma^{(k)}(\mathbf{x}_i^{(k-1)}, \Pi_j \lambda^{(k)}(\mathbf{x}_i^{(k-1)}, \mathbf{x}_j^{(k-1)}, \mathbf{e}_{ij}))$$

where j ∈ 𝒩(i) denotes a neighbor node of node i. $\mathbf{x}_i$ is the node feature vector and $\mathbf{e}_{ij}$ is the edge feature vector. γ and λ denote differentiable functions and Π denotes a differentiable aggregation function.

Specifically, we use GCNs [4] which implement message passing using the adjacency matrix **A**:

$$\mathbf{X}^{(k)} = \check{\mathbf{D}}^{-1/2}\check{\mathbf{A}}\check{\mathbf{D}}^{-1/2}\mathbf{X}^{(k-1)}\boldsymbol{\Omega},$$

where $\check{\mathbf{A}} = \mathbf{A} + \mathbf{I}$ denotes the adjacency matrix with inserted self-loops and $\check{D}_{ii}=\sum_{j=0}\check{A}_{ij}$ its diagonal degree matrix. $A_{ij}$ is one when there is an edge from node i to node j, and zero when there is no edge. Thus, all edges have a *weight of one*.

A graph in PyG is described by an instance of the *Data* class, which has the following attributes: *x* (node feature matrix), *edge_index* (edge indices), *edge_attr* (edge feature matrix), *pos* (node position matrix) and *y* (target). The PyG implementation of GCN, *GCNConv*, employs the *edge_weight* vector in graph convolutions. However, in *standard (default) graph convolutions*, the edge_weight vector has values of one, i.e., all edges have a *weight of one*.

## 5.2 Geometric Graph Convolutions

For proof of concept, we use the *distance-geometric graph representation* for geometric graph convolutions. This representation is based on distances: *edge distances (d), angle distances ($d^\theta$) and dihedral distances ($d^\varphi$)*, as discussed in Section 4.3. Further, to utilize standard GCNs for geometric graph convolutions, we employ a simple *edge weight / edge distance correlations* scheme whose parameters can be fixed using *reference values* or determined using *Bayesian hyperparameter optimization*.

The combination of using the distance-geometric graph representation and employing an edge weight / edge distance correlations scheme enables us to incorporate the full geometry of 3D graphs in graph convolutions utilizing standard GCNs by (1) expanding the kinds of edges involved to include not just *edges (e)* with neighbor nodes, but also *angle edges ($e^\theta$)* with second-neighbor nodes and *dihedral edges ($e^\varphi$)* with third-neighbor nodes and (2) assigning different *weight*s to different edges based on their kind and their distance.

The first extension to standard GCNs means that in Section 5.1 we now have $j \in \mathcal{N}(i) + \mathcal{N}^\theta(i) + \mathcal{N}^\varphi(i)$, with $\mathcal{N}(i)$ being the first neighbors, $\mathcal{N}^\theta(i)$ the second neighbors and $\mathcal{N}^\varphi(i)$ the third neighbors, of node i. The second extension is discussed below.

## 5.3 Edge Weight / Edge Distance Correlations

We employ a simple *edge weight / edge distance correlations* scheme whose parameters can be fixed using *reference values* or determined using *Bayesian hyperparameter optimization*, with the latter to be discussed in the next section.

We recall that, as discussed in Section 2.1, for molecular graphs, bond strength (edge weight) is empirically related to bond length (edge distance) through power laws $s = (R/R_o)^{-N}$ with parameters $R_0$ and $N$. We leverage this knowledge and correlate edge weight with edge distance through *power laws with parameters $R_0$ and $N$*. However, this is not necessary in general. For other types of 3D graphs, even molecular graphs, other forms of edge weight / edge distance correlations can be used, e.g., exponential functions.

For *edges (e)*, we define the edge weight (w) / edge distance (d) correlation as follows:

$$w = (d / R_0)^{-N}$$

For *angle edges ($e^\theta$)*, we define the edge weight ($w^\theta$) / edge distance ($d^\theta$) correlation as:

$$w^\theta = (d^\theta / R^\theta_0)^{-N\theta}$$

And for *dihedral edges ($e^\varphi$)*, we define the edge weight ($w^\varphi$) / edge distance ($d^\varphi$) correlation as:

$$w^\varphi = (d^\varphi / R^\varphi_0)^{-N\varphi}$$

The variation of edge weight w.r.t. edge distance, as a function of $R_0$ and $N$, is shown in the following table:

| Edge distance / $R_0$ | Edge weight (N=2) | Edge weight (N=3) | Edge weight (N=4) | Edge weight (N=5) | Edge weight (N=6) |
|---|---|---|---|---|---|
| 1 | 1 | 1 | 1 | 1 | 1 |
| 1.5 | 0.444 | 0.296 | 0.198 | 0.132 | 0.088 |
| 2 | 0.25 | 0.125 | 0.063 | 0.031 | 0.016 |
| 2.5 | 0.160 | 0.064 | 0.026 | 0.010 | 0.004 |
| 3 | 0.111 | 0.037 | 0.012 | 0.004 | 0.001 |

In contrast to standard graph convolutions which use edge_weight with values of one, *geometric graph convolutions* use edge_weight with values calculated from edge distances in GCNConc (see Section 5.1)

## 5.4 Bayesian Hyperparameter Optimization

If no reference values are available, we can treat the parameters $(R_0, N)$, $(R^\theta_0, N^\theta)$ and $(R^\varphi_0, N^\varphi)$ as hyperparameters and use *Bayesian hyperparameter optimization* [2] to find the best values. *Bayesian optimization* [19-23] is a powerful tool for the joint optimization of hyperparameters, efficiently trading off exploration and exploitation of the hyperparameter space.

Following [2], we implement Bayesian hyperparameter optimization using *BoTorch, GPyTorch* and *Ax* [24-30]. Ax provides the *optimize*() function to construct and run a full *OptimizationLoop*, and to return the *best hyperparameter configuration*. We call the optimize() function with the following input parameters:

- hyperparameters: $R_0$, $N$, $R^\theta_0$, $N^\theta$, $R^\varphi_0$, $N^\varphi$
- evaluation function: train_evaluate
- minimize: True
- total trials (default: 20)

The evaluation function, *train_evaluate(parametrization)*, that we provide has three components:

1. model: geometric graph convolution (GGC)
2. train(training dataset, parametrization)
3. evaluate(validation dataset, parametrization)

The *hyperparameter configuration* (parametrization) is automatically generated by Ax for each Trial during a full run of the OptimizationLoop.

For each Trial, given a hyperparameter configuration, the *train()* function trains the GGC model (which employs GCNConv) using the training dataset. Once the GGC model is trained, the *evaluate()* function evaluates the model using the validation dataset and returns the *RMSE (Root Mean Squares Error)* which serves as the *objective score* for use in optimization.

## 6 Experiments

For the feasibility study, we carried out a number of experiments using the *ESOL* and *FreeSolv* datasets, which are used in [5-6] for training and evaluating 3D-extended GCNs. In particular, we used the dataset files provided by Geo-GCN [6], which contain molecular graph data including three-dimensional node coordinates. These are small datasets with *901 / 113 / 113* and *510 / 65 / 64* training / test / validation samples, respectively. Our focus, however, is on qualitatively comparing results of geometric graph convolutions with those of standard graph convolutions, all based on the same sample sizes. That is, our interest is on relative accuracy not absolute accuracy.

The results are listed in the following table. *Standard GC* (graph convolutions) utilizes the default GCNConv with all edges having a weight of one, discussed in Section 5.1, and serves as the *baseline* for comparison. *Geometric GC* utilizes the GCNConv with edge weights calculated from edge distances, discussed in Section 5.3.

*Geometric GC (Ref)* denotes the reference geometric GC which uses fixed $R_0$ = 1.39 and N = 4.55 (i.e., 1/0.22), discussed in Sections 2.1. This applies to all three cases of geometric graph convolutions: edges (*1st Nbrs*) which include first-neighbor nodes, edges + angle edges (*2nd Nbrs*) which include first- and second-neighbor nodes, and edges + angle edges + dihedral edges (*3rd Nbrs*) which include first-, second- and third-neighbor nodes.

*Geometric GC (BHO)* represents the geometric GC which uses the best ($R_0$, N)'s obtained through Bayesian hyperparameter optimization, discussed in Section 5.4. We adopted the following ranges for the hyperparameters: $R_0$ (in Å) and $N$ in the range of 1.0 to 3.0 and 2.0 to 6.0, respectively, as discussed in Section 2.1; $R^\theta_0$ (in Å) and $N^\theta$ in the range of 1.0 to 3.0 (discussed in Section 3.1) and 2.0 to 6.0, respectively; $R^\varphi_0$ (in Å) and $N^\varphi$ in the range of 1.0 to 4.0 (discussed in Section 3.1) and 2.0 to 6.0, respectively. For the study, we employed 40 trials and, for each trial, 50 epochs. This applies to all three cases of geometric graph convolutions: edges (*1st Nbrs*), edges + angle edges (*2nd Nbrs*), and edges + angle edges + dihedral edges (*3rd Nbrs*).

We include the 1st Nbrs and 2nd Nbrs cases to verify and show the consistency of Geometric GC results. The (full) geometric graph convolutions are represented by the *3rd Nbrs* cases.

| Dataset | Model | $R_0$ (Å) | N | $R^\theta_0$ (Å) | $N^\theta$ | $R^\phi_0$ (Å) | $N^\phi$ | RMSE |
|---|---|---|---|---|---|---|---|---|
| ESOL | Standard GC | - | - | - | - | - | - | 0.4573 |
| | Geometric GC (Ref) – 1$^{st}$ Nbrs | 1.39 | 4.55 | - | - | - | - | 0.4283 |
| | Geometric GC (BHO) – 1$^{st}$ Nbrs | 1.0 | 2.6447 | - | - | - | - | 0.4163 |
| | Geometric GC (Ref) – 2$^{nd}$ Nbrs | 1.39 | 4.55 | 1.39 | 4.55 | - | - | 0.4275 |
| | Geometric GC (BHO) – 2$^{nd}$ Nbrs | 3.0 | 4.7603 | 1.4402 | 3.7508 | - | - | 0.4214 |
| | Geometric GC (Ref) – 3$^{rd}$ Nbrs | 1.39 | 4.55 | 1.39 | 4.55 | 1.39 | 4.55 | 0.4273 |
| | Geometric GC (BHO) – 3$^{rd}$ Nbrs | 1.3385 | 4.2756 | 1.6828 | 5.4836 | 1.8408 | 6.0 | 0.4261 |
| FreeSolv | Standard GC | - | - | - | - | - | - | 0.4183 |
| | Geometric GC (Ref) – 1$^{st}$ Nbrs | 1.39 | 4.55 | - | - | - | - | 0.3706 |
| | Geometric GC (BHO) – 1$^{st}$ Nbrs | 1.5783 | 4.9247 | - | - | - | - | 0.3753 |
| | Geometric GC (Ref) – 2$^{nd}$ Nbrs | 1.39 | 4.55 | 1.39 | 4.55 | - | - | 0.3711 |
| | Geometric GC (BHO) – 2$^{nd}$ Nbrs | 2.6367 | 3.3691 | 1.5782 | 3.000 | - | - | 0.3745 |
| | Geometric GC (Ref) – 3$^{rd}$ Nbrs | 1.39 | 4.55 | 1.39 | 4.55 | 1.39 | 4.55 | 0.3710 |
| | Geometric GC (BHO) – 3$^{rd}$ Nbrs | 1.5146 | 3.8488 | 2.5738 | 3.3665 | 3.1133 | 2.4705 | 0.3764 |

It can be seen from the table, for both the ESOL and FreeSolv datasets, the results of *Geometric GC* show significant improvement over those of Standard GC. They demonstrate the importance of incorporating *geometry, using the distance-geometric graph representation,* in deep learning on 3D graphs.

## 7 Summary and Conclusion

The geometry of three-dimensional (3D) graphs, consisting of nodes and edges, plays a crucial role in many important applications. To facilitate the incorporation of geometry in deep learning on 3D graphs, we defined three types of geometric graph representations: positional, angle-geometric and distance-geometric.

For proof of concept, we used the distance-geometric graph representation for geometric graph convolutions. Further, to utilize standard graph convolution networks, we employed a simple edge weight / edge distance correlation scheme, whose parameters can be fixed using reference values or determined through Bayesian hyperparameter optimization. The results of geometric graph convolutions, for the ESOL and Freesol datasets, showed significant improvement over those of standard graph convolutions.

Our work demonstrates the feasibility and promise of incorporating geometry, using the distance-geometric graph representation, in deep learning on 3D graphs.

**Acknowledgement:** Thanks to my wife Hedy (期芳) for her support.